\documentclass[conference]{IEEEtran}
\IEEEoverridecommandlockouts
\usepackage{cite}
\usepackage{amsmath,amssymb,amsfonts}
\usepackage{algorithmic}
\usepackage{graphicx}
\usepackage{textcomp}
\usepackage{xcolor}
\usepackage{times}  % DO NOT CHANGE THIS
\usepackage{helvet}  % DO NOT CHANGE THIS
\usepackage{courier}  % DO NOT CHANGE THIS
\usepackage[hyphens]{url}  % DO NOT CHANGE THIS
\usepackage{graphicx} % DO NOT CHANGE THIS
\usepackage[square,sort,comma,numbers]{natbib}
\usepackage{caption} % DO NOT CHANGE THIS AND DO NOT ADD ANY OPTIONS TO IT
\DeclareCaptionStyle{ruled}{labelfont=normalfont,labelsep=colon,strut=off} % DO NOT CHANGE THIS
\usepackage{tikz}

\usepackage{algorithmic}
\usepackage{algorithm}
\usepackage[algo2e,boxed,ruled]{algorithm2e} 
\SetKwRepeat{Do}{do}{while}
\usepackage{amsthm} 
\usepackage{mathtools}
\usepackage{booktabs}

\usepackage{longtable}
\usepackage[format=plain,indention=0cm, font=small, labelfont=bf]{caption}
\usepackage{tabu}
\usepackage{amsmath}
\usepackage{amssymb}
\usepackage{subfloat}

\usepackage[utf8]{inputenc} % allow utf-8 input
\usepackage[T1]{fontenc}    % use 8-bit T1 fonts
\usepackage{hyperref}       % hyperlinks
\usepackage{url}            % simple URL typesetting
\usepackage{booktabs}       % professional-quality tables
\usepackage{amsfonts}       % blackboard math symbols
\usepackage{nicefrac}       % compact symbols for 1/2, etc.
\usepackage{microtype}      % microtypography
\usepackage{xcolor}         % colors
\usepackage{microtype}
\usepackage{amsmath}

\usepackage{times}
\usepackage{latexsym}

\usepackage{graphicx}
\usepackage{amsmath}
\usepackage{algorithmic}
\usepackage{algorithm}
\usepackage{subfig}
\usepackage{microtype}
\usepackage{color}

\usepackage{xurl}
\usepackage{multirow}
\usepackage{microtype}
\usepackage{color}
\usepackage{microtype}
\usepackage{subfloat}

\usepackage[utf8]{inputenc}

\usepackage{sidecap}
\usepackage{times}
\usepackage{latexsym}
\usepackage{tikz}
\usepackage{graphicx}
\usepackage{bm}

\usepackage{amsmath}
\usepackage{algorithmic}
\usepackage{algorithm}
\usepackage{booktabs}

\usepackage{multirow}

\newtheorem{proposition}{Proposition}

\newtheorem{remark}{Remark}
\newtheorem{assumption}{Assumption}
\usepackage{newfloat}
\usepackage{listings}
\floatstyle{ruled}
\newfloat{listing}{tb}{lst}{}
\floatname{listing}{Listing}

\usepackage{bibentry}

\usepackage[draft]{todonotes}   % notes showed

\def\BibTeX{{\rm B\kern-.05em{\sc i\kern-.025em b}\kern-.08em
    T\kern-.1667em\lower.7ex\hbox{E}\kern-.125emX}}
\begin{document}

% \title{Progressive Sparse Training on Spiking Neural Network*\\
\title{Dynamic Sparse Training via Balancing the Exploration-Exploitation Trade-off}

\author{
    Shaoyi Huang  \textsuperscript{\rm 1},
    Bowen Lei \textsuperscript{\rm 2},
    Dongkuan Xu \textsuperscript{\rm 3}, 
    Hongwu Peng \textsuperscript{\rm 1},
    Yue Sun \textsuperscript{\rm 4},
    Mimi Xie \textsuperscript{\rm 5},
    Caiwen Ding \textsuperscript{\rm 1}\\
 %\affiliations
 $^1$University of Connecticut,
 $^2$Texas A\&M University,
 $^3$North Carolina State University,\\
 $^4$Lehigh University,
 $^5$University of Texas at San Antonio\\
 \small \{shaoyi.huang, hongwu.peng, caiwen.ding\}@uconn.edu,\\
 \small bowenlei@stat.tamu.edu, 
 dxu27@ncsu.edu,
 yus516@lehigh.edu,
 mimi.xie@utsa.edu
 }
 
% \renewcommand{\shortauthors}{Huang, et al.}

% \and
% \and
% \and
% \and
% \and

\maketitle

\begin{abstract}
Over-parameterization of deep neural networks (DNNs) has shown high prediction accuracy for many applications. Although effective, the large number of parameters hinders its popularity on resource-limited devices and has an outsize environmental impact. Sparse training (using a fixed number of nonzero weights in each iteration) could significantly mitigate the training costs by reducing the model size. However, existing sparse training methods mainly use either random-based or greedy-based drop-and-grow strategies, resulting in local minimal and low accuracy.
% As a result, exploring sparse end-to-end training has emerging as an important research trend. 
In this work, to assist explainable sparse training, we propose important weights \underline{E}xploitation and coverage \underline{E}xploration to characterize \underline{D}ynamic \underline{S}parse \underline{T}raining (DST-EE), and provide quantitative analysis of these two metrics. We further design an acquisition function and provide the theoretical guarantees for the proposed method and clarify its convergence property.
% In particular, we justify the influence of the mask-induced error.
Experimental results show that sparse models (up to 98\% sparsity) obtained by our proposed method outperform the SOTA sparse training methods %algorithms 
on a wide variety of deep learning tasks. On VGG-19 / CIFAR-100, ResNet-50 / CIFAR-10, ResNet-50 / CIFAR-100, our method has even higher accuracy than dense models. On ResNet-50 / ImageNet, the proposed method has up to 8.2\% accuracy improvement compared to SOTA sparse training methods.
% we consider the dynamic sparse training as a sparse connectivity search problem and design an exploitation and exploration acquisition function to escape from local optima and saddle points. A series of experimental results show the proposed method could find better connectivity and achieve the state-of-the-art accuracy with ResNet-18 on CIFAR10, ResNet-50 on ImageNet.
\end{abstract}


\section{Introduction}
% Deep neural networks (DNNs) have made significant progress and great success in a variety of real-world scenarios over the last few decades.
Increasing deep neural networks (DNNs) model size has shown superior prediction accuracy in a variety of real-world scenarios
% have a dominant effect on the tremendous performance
\cite{liu2021we}. However, as model sizes continue to scale, a large amount of computation and heavy memory requirements prohibit the DNN training on resource-limited devices, as well as being environmentally unfriendly~\cite{huang2022sparse, peng2022length, qi2021accommodating, peng2022towards, qi2021accelerating, manu2021co}. 
{A Google study showed that GPT-3~\cite{gpt3} (175 billion parameters) consumed 1,287 MWh of electricity during training and produced 552 tons of carbon emissions, equivalent to the emissions of a car for 120 years~\cite{patterson2021carbon}.} Fortunately, {\textit{sparse training could significantly mitigate the training costs by using a fixed and small number of nonzero weights in each iteration,
% reducing the model size and the number of computations,
while preserving the prediction accuracy for downstream tasks.}}

\begin{figure}[!h]
    \centering
    % \vspace{-.25in}
    \subfloat[Non-active weights with small initial gradients are ignored in greedy-based weight growth methods (i.e., RigL, ITOP, ...)]{\label{fig:motivation1}
    \hspace{-.0in}\includegraphics[width=0.95\columnwidth]{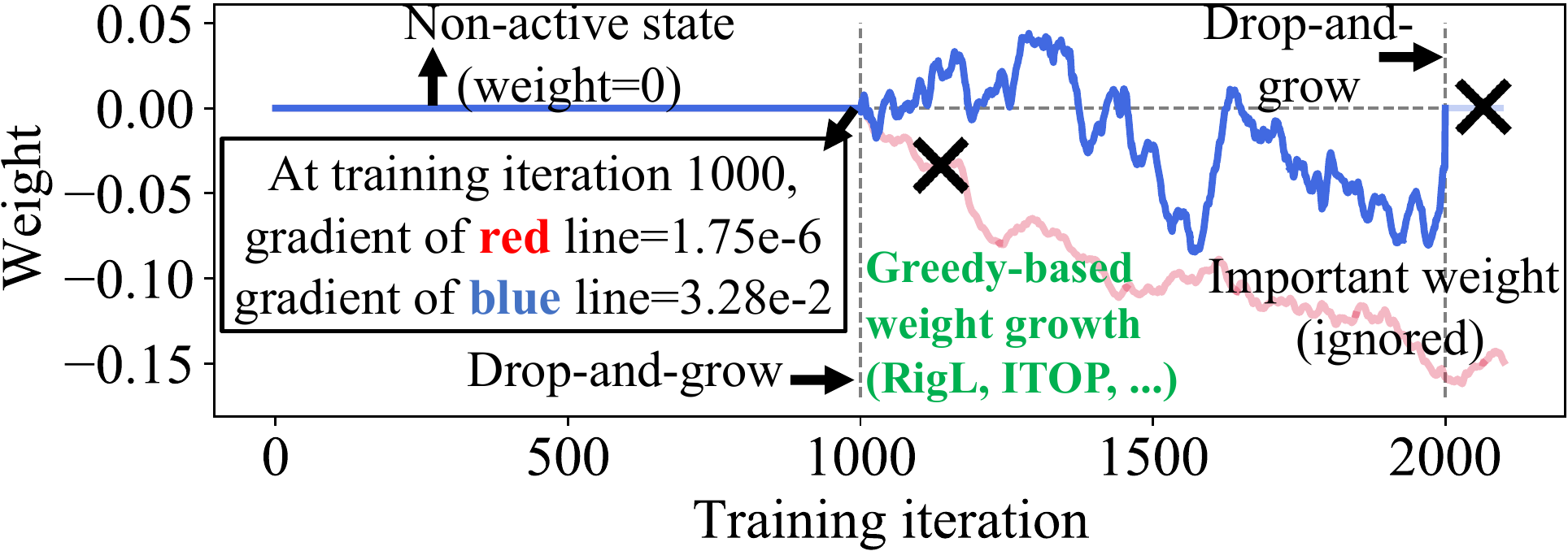}}
    %\caption{fig1}
    % \vspace{-.0in}
    \quad
    \subfloat[
    % Our method could grow weights that have never been explored before. 
    Non-active weights with small initial gradients could be retained and grown in proposed method.]{\label{fig:motivation2}
    \hspace{-.0in}\includegraphics[width=1.\columnwidth]{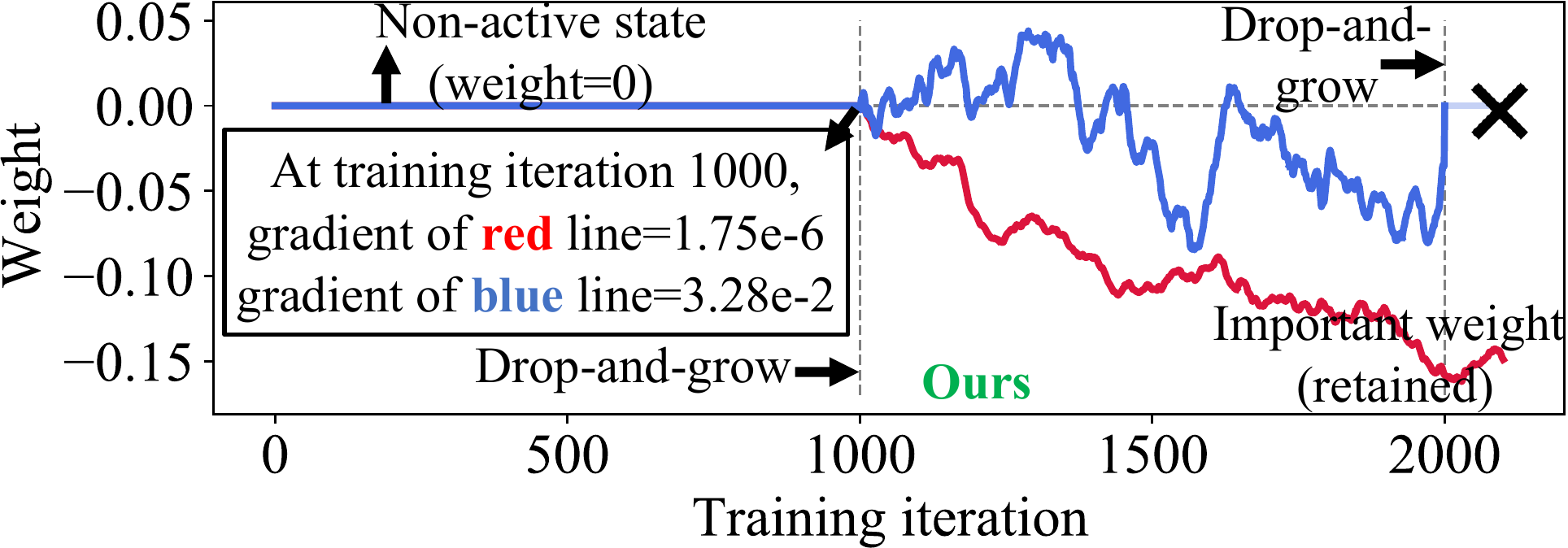}}
    \caption{Gradient-based weight growth methods vs. proposed method. (a) The red line shows the weight with a small gradient is ignored (not grown), while the blue line denotes that the weight with a large gradient is grown at iteration=1000. (b) Weight with a small gradient at iteration=1000 can be grown applying our method, and at training iteration = 2000 it is more important.
    % than the weight in blue.
    }
    \label{fig:motivation}
    % \vspace{0.2in}
\end{figure}

% X-axis is the number of training step and y-axis is weight value. Black dashed lines denote prune-and-grow in corresponding step. Unimportant (pruned) weight with large gradient (blue curve) vs. important (retained) weight with small gradient (red curve) (b) Percentage of weights ignored by gradient magnitude growth methods (i.e., RigL, ITOP) on VGG-19/CIFAT-100. The weights have small initial gradients (red line at iteration=2000 in (a)). X-axis is the selected layer name, for example, "f3" means layer feature.3.weight.

Two research trends on sparse training have attracted enormous popularity. One is~\textit{static mask}-based method~\cite{lee2019snip,wang2020picking, peng2022towards},
% e.g., SNIP~\cite{lee2019snip}, GraSP~\cite{wang2020picking}, 
where sparsification starts 
% given network 
at initialization before training. Afterward, the sparse mask (a binary tensor corresponding to the weight tensor) is fixed. Such limited flexibility of subnetwork or mask selection
% is limited 
leads to sub-optimal subnetworks with poor accuracy. 
% non-zero weight elements are updated in a standard way
% After initialization, the non-zero weight elements are updated in a standard way, while the zero elements are kept at zero. 
% Although static mask training reduces the memory and computation demands of training resources, the sparse structure of network is fixed at initialization and the flexibility of subnetwork selection is limited, leading to sub-optimal subnetworks. 
To improve the flexibility, \textit{dynamic mask training} has been proposed~\cite{mocanu2018scalable,mostafa2019parameter,evci2020rigging}, where the sparse mask is periodically updated by drop-and-grow to search for better subnetworks with high accuracy, where in the drop process we deactivate a portion of weights from active states (nonzero) to non-active states (zero), vice versa for the growing process. 
% while grow means activatinge same number of weights from non-active states to active ones.
% In the above dynamic sparse training methods, DeepR~\cite{bellec2017deep}, SET~\cite{mocanu2018scalable}, and DSR~\cite{mostafa2019parameter} explored new weights randomly, which leads to low accuracy. While RigL~\cite{evci2020rigging} and  ITOP~\cite{liu2021we} greedily grow the weights with the high gradient magnitude, leading to limited coverage of the weights, which results in sub-optimal subnetworks.
% However, we observe that these methods mainly greedily grow the weights with the high gradient magnitude, \textcolor{red}{(CD:need a better summary of the disadvantage of existing sparse training method)} resulting in limited coverage of the weights. 

However, these methods mainly use either random-based or greedy-based growth strategies. The former one usually leads to lower accuracy while the latter one greedily searches for sparse masks with a local minimal in a short distance~\cite{he2022sparse}, resulting in limited weights coverage and thus a sub-optimal sparse model. As an illustration in Figure~\ref{fig:motivation1} using VGG-19/CIFAR-100, at one drop-and-grow stage (1,000th iteration), the gradient-based approach grows non-active weights with relatively large gradients but ignores small gradients. However, as training continues (e.g., at the 2,000th iteration), these non-active weights with small gradients will have large magnitude and hence are important to model accuracy~\cite{renda2020comparing,zafrir2021prune}. Therefore, they should be considered for the growth at the 1,000th iteration as shown in Figure~\ref{fig:motivation2}.
% non-active weights with large gradient magnitude (blue line) for growthon VGG-19/CIFAR-100,
In addition, more than 90\% of non-active weights but important weights are ignored in 12 out of 16 convolutional layers.
To better preserve these non-active weights but important weights, we propose a novel weights \underline{E}xploitation and coverage \underline{E}xploration characterized \underline{D}ynamic \underline{S}parse \underline{T}raining (DST-EE) to update the sparse mask and search for the “best possible” subnetwork. Different from existing greedy-based methods, which only exploit the current knowledge, we further explore and grow the weights that have never been covered in past training iterations,
% (i.e., we preferentially grow weights that have never been covered before), 
thus increasing the coverage of weights and avoiding the subnetwork searching process being trapped in a local optimum~\cite{li2015ps}.
% Specifically, on the one hand, we exploit the current knowledge (growing the magnitude gradients in the next step, thus rapidly reducing the loss), and one the other hand, we explore the weights that have never been covered in past training steps (i.e., we preferentially grow weights that have never been covered before), thus increasing the coverage of weights and avoiding the subnetwork searching process being trapped in a local optimum~\cite{li2015ps}.
% \textcolor{red}{underline our contribution; in contrast/different from existing method}
% We carefully designed an acquisition function which calculates the importance score of each growth candidate using the combination of the gradient tensor and the occurrence counter tensor. Every $\Delta T$ iterations, weights with least magnitude values are set to zeros, and weights with high importance score will be grown to maintain the sparsity of the weight tensor. This process is repeated till the end of training, which never requires intensive dense model training cost during training, thus significantly reducing memory footprints proportionally to the network sparsity.
The contributions of the paper are summarized as follows:

\begin{itemize}

\item To assist explainable sparse training, we propose \textbf{\textit{important weights exploitation}} and \textbf{\textit{weights coverage exploration}} to characterize sparse training. We further provide the quantitative analysis of the strategy and show the advantage of the proposed method.

\item We design an acquisition function for the growth process. We provide theoretical analysis for the proposed exploitation and exploration method and clarify the convergence property of the proposed sparse training method.

% In particular, we justify the influence of the mask-induced error. 
% Our exploration method is guided by a rigorous theoretical analysis. 

% \item Our proposed method does not need to train dense models throughout the training process, achieving up to 95\% sparsity ratio and even higher accuracy than dense training, with same amount of iterations.

\item Our proposed method does not need to train dense models throughout the training process, achieving up to 95\% sparsity ratio and even higher accuracy than dense training, with same amount of iterations.  Sparse models obtained by the proposed method outperform the SOTA sparse training methods.
% compared to SOTA sparse training methods.
% \textcolor{red}{We further show the scalability of proposed method on graph neural network (GNN). On both wiki-talk and ia-email datasets, our method even outperforms prune-from-dense using admm algorithm, achieving up to 23.3\% higher link prediction accuracy.} \textcolor{red}{(where are GNN results?)}

% on ResNet-50 / CIFAR-10,  ResNet-50 / CIFAR-100, VGG-19 / CIFAR-100
% with the same number of training epochs,
% greatly reducing the training memory footprints.

% \item We use prune-and-grow which has an advantage in computation over
% grow-and-prune. For example, \textcolor{red}{ResNet-50 XXX}

\end{itemize}

On VGG-19 / CIFAR-100, ResNet-50 / CIFAR-10, ResNet-50 / CIFAR-100, our method has even higher accuracy than dense models. On ResNet-50 / ImageNet, the proposed method has up to 8.2\% accuracy improvement. On graph neural network (GNN), our method
% even 
outperforms prune-from-dense using ADMM algorithm~\cite{peng2021accelerating, chen2021re, zhang2018systematic}, achieving up to 23.3\% higher link prediction accuracy.

\section{Related Work}

\begin{figure*}[!h]
\begin{center}
    %  \vspace{-.1in}
 \includegraphics[width = 0.97\linewidth]{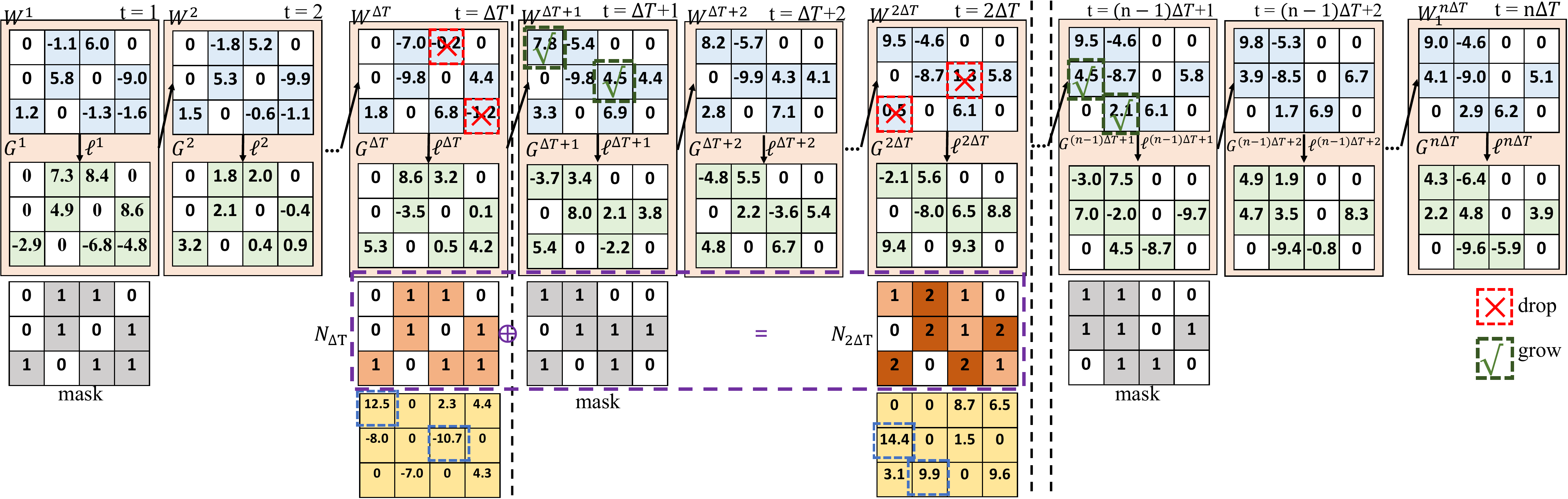}
     \caption{Sparse training data flow of proposed method.}
     \vspace{-3mm}
     \label{fig:dataflow}
\end{center}
\end{figure*}

% \textbf{Dynamic Mask Training.} 
% Under the fixed parameter count,
% Dynamic masks training adaptively updates mask tensors and provides more flexibility and hence higher accuracy.
% , given  the mask could be updated / fixed parameter count,
% fixed computational cost throughout training and index/mask flexible -> high performance
Sparse Evolutionary Training (SET)~\cite{mocanu2018scalable} removed least
magnitude valued weights and randomly grow the corresponding number of weights back at the end of each training epoch. SNFS~\cite{dettmers2019sparse} utilized exponentially smoothed momentum to find the important weights and layers, and redistributed pruned weights based on the mean momentum magnitude per layer. 
RigL~\cite{evci2020rigging} updated the sparsity topology of the sparse network during training using the same magnitude-based weights dropping method while growing back the weights using top-k absolute largest gradients, achieving better accuracy than static mask training under same sparsity. However, the greedy-based growth policy leading to limited weights coverage, therefore a sub-optimal sparse model.
% It achieved comparable accuracy to static mask training methods, but only
% using sparse neural networks with a fixed parameter count and a fixed computational cost throughout training; 
ITOP~\cite{liu2021we} discovered that the benefits of dynamic mask training come from its ability to consider across time all possible parameters. 
% However, all ITOP, SNFS and RigL need to compute the dense gradient related to both pruning and unpruned weights. Therefore, they are, essentially, not computational and memory efficient. 
In addition, MEST~\cite{yuan2021mest} employed a gradually decreasing drop and grow rate with a more relaxed range of parameters for growing. However, both ITOP and MEST keep the same drop-and-growth strategy as the existing works and have limited weights coverage.  GaP~\cite{ma2021effective} divides the DNN into several partitions, growing one partition at a time to dense and pruning the previous dense partition to sparse, with the aim of covering all weights. However, it requires more training time than traditional pruning methods, which limits its application on resources limited scenarios.

\section{Important
weights exploitation and  coverage exploration}

\subsection{Overview}

% Figure~\ref{fig:workflow} is a toy example of sparse training of a two-layer convolutional neural network using proposed method. The two layers have corresponding two weight tensors $W_1$ and $W_2$, two gradient tensors $G_1$ and $G_2$, two occurrence frequency counter tensors $N_1$ and $N_2$, two importance score tensors $S_1$ and $S_2$, two growth candidate tensors $C_1$ and $C_2$. The neural network is initialized at 0.5 sparsity. After initialization, following actions will be applies on the model: \protect\circled{1} train model $\Delta T - 1$ iterations by updating nonzero weights; \protect\circled{2} drop 2 least magnitude weights (set as 0) at the $\Delta T$ iteration; \protect\circled{3} grow 2 weights back which has the largest importance scores calculated using Equation~\ref{eq:ee}. The iterations \protect\circled{1} - \protect\circled{4} will be repeated in the remaining training iterations. 

We formalize the sparse training process of the proposed DST-EE as follows. We define a $L$-layer deep neural network with dense weight $\bf W = {[\bf W_1, \bf W_2, ..., \bf W_L]}$. During the training process, the weight of $i$-th layer at $t$-th iteration is denoted by ${\textbf{W}}_{i}^t$. We randomly initialize sparse weight tensor as ${\textbf{W}'}$ = ${[\bf {W}'_1, \bf {W}'_2, ..., \bf {W}'_L]}$ with sparsity distribution of $P$ using ERK~\cite{mocanu2018scalable} initialization. Each sparse weight tensor within a layer has a corresponding mask tensor (zero elements masked by 0 and other elements masked by 1) with the same size.
% \textbf{Initialization}. We define a deep neural network training progress, 
% %  with $N$ layers indexed by $n \in 1,...,N$,
% where the weights at iteration $t$ are denoted by ${{\bf W}}^{t}$. We randomly initialize sparse weight tensor as $\mathbf{W}^0$ with sparsity of $\theta_0$ using ERK~\cite{mocanu2018scalable} initialization. Each sparse weight tensor within a layer has a corresponding mask tensor (zero elements masked by 0 and other elements masked by 1) with same size. 
% , and compute the sparse training masks based on the magnitude.
% \textbf{Training}. 
We define zero elements in weight tensor as non-active weights and others as active weights. For each iteration, we only update the active weights.
% We update the mask tensor every $\Delta T$ iterations.
%  
% Every $\Delta T$ iterations, we update the mask tensor, and substitute the initialized weights $W_0$ accordingly.
% \textbf{Drop (deactivate):} 
In addition, every $\Delta T$ iteration, we update the mask tensor, i.e., for $i$-th layer, we 
% compute the $l2$-norm of each kernel and 
drop the $k_i$ weights that are closest to zero (i.e., smallest positive weights and the largest negative weights), the dropped weights are denoted by ArgTopK$(\textbf{W}'_i, k_i)$.
% we compute the $l2$-norm of each kernel. We keep the top-$k$ elements and drop the rest. 
% We denote $\{\textbf{W}_j\}_{j=1}^{j=i}$ as the collection of weights in the first $i$ layers
% $\textbf{N}_{\textbf{W}}^{t}$
% \begin{figure*}[!h]
% \begin{center}
%     %  \vspace{-.1in}
%  \includegraphics[width = 0.97\linewidth]{figures/data flow cropped.pdf}
%      \caption{Sparse training data flow{(XXX)}.}
%     %  \vspace{-3mm}
%      \label{fig:dataflow}
% \end{center}
% \end{figure*}
% \textbf{Grow (activate):} 
We denote $\textbf{N}_{i}^{t}$ as the counter tensor that collects the occurrence frequency for each 1 mask. We initialize $\textbf{N}_{i}^{t}$ as a zero tensor with the same size as the corresponding weight tensor. Every $\Delta T$ iteration, the counter tensor is updated by adding the counter tensor with the existing mask tensor. We use ${\textbf{S}}_i^{t}$ to denote the importance score tensor in $q$-th mask update. We design the following acquisition function to compute the importance score tensor

\begin{equation}
\begin{aligned}
\small
{\textbf{S}_i^{t} = |\frac {\partial l(\textbf{W}_i^{t}, \mathcal{X})}{\partial \textbf{W}_i^{t}}| + c \frac{\ln t}{\textbf{N}_i^{t} + \epsilon},\quad t = q \Delta T,\quad i = 1, 2, ..., L}
\label{eq:ee} 
\end{aligned}
\end{equation}
% where the addition is composed of two terms.
where the first term $|\frac {\partial l(\textbf{W}_i^{t}, \mathcal{X})}{\partial \textbf{W}_i^{t}}|$ is the absolute gradient tensor of $i$-th layer at $t$-th iteration. $\partial l(\textbf{W}_i^{t}, \mathcal{X})$ is the loss of $i$-th layer.  $\mathcal{X}$ is the input training data. 
In the second term $c \frac{\ln t}{\textbf{N}_i^{t} + \epsilon}$, $c$ is the coefficient to balance between the two terms and $\epsilon$ is a positive constant to make the remainder as nonzero. For each importance score tensor, we identify the $k$ highest absolute values and select the indices. These corresponding mask values with the same indices will be set to 1s. In the next iteration, we update the weights using the new mask tensor. In the whole process, we maintain that the newly activated weights are the same amount as the previously deactivated weights. We repeat the aforementioned iterations till the end of training. The details of our method are illustrated in Algorithm~\ref{alg:dst-ee}, where $\cdot$ means tensor matrix multiplication.

\begin{algorithm}[!h]
\footnotesize
  \caption{DST-EE}
  \label{alg:dst-ee}
\begin{algorithmic}
    \STATE {\bfseries Input:} a $L$-layer network $f$ with dense weight $\textbf{W} = {\textbf{W}_1, \textbf{W}_2,  ..., \textbf{W}_L}$; sparsity distribution: $P = {P_1, P_2, ..., P_L}$; total number of training iterations $T_{end}$.
    
    \STATE {\bfseries Set} $\mathcal{X}$ as the training dataset; $\Delta T$ as the update frequency; $\alpha$ as the learning rate; ${k_1, k_2, ..., k_L}$ are variables denoting the number of weights dropped every $\Delta T$ iterations; ${\textbf{M}_1, \textbf{M}_2,  ..., \textbf{M}_L}$ are the sparse masks. ${\textbf{S}_1, \textbf{S}_2,  ..., \textbf{S}_L}$ are the importance score tensors.
    
    \STATE {\bfseries Output:} a $L$-layer sparse network with sparsity distribution $P$.
  
  \STATE $\textbf{W}'={\textbf{W}'_1, \textbf{W}'_2,  ..., \textbf{W}'_L} \leftarrow$ sparsify ${\textbf{W}_1, \textbf{W}_2,  ..., \textbf{W}_L}$ with $P$
  
  \STATE $\textbf{N}_i^{t} \leftarrow \textbf{M}_i$
    
    \FOR{each training iteration $t$ }
        \STATE Loss $\delta_t$ $\leftarrow$ $f(x_t,\textbf{W}')$, $x_t \in \mathcal{X}$
        \IF {$t$ (mod $\Delta T$) == 0 and $t$ $<$ $T_{end}$}
            \FOR{$0 < i < L + 1$}
                \STATE $\textbf{W}'_i \leftarrow$ ArgDrop$(\textbf{W}'_i, $ArgTopK$(\textbf{W}'_i, k_i))$
                % \STATE $S_{exploi}(i) = \nabla(W'_i) \delta_t$
                % \STATE  $S_{explor}(i) = \frac{\ln t}{N_{W'_i}^{t} + \epsilon}$
                % \STATE  $S_{total} = S_{exploi}(i) + \lambda * S_{explor}(i)$
                \STATE  $\textbf{S}_i = \nabla(\textbf{W}'_i) \delta_t + c * \frac{\ln t}{\textbf{N}_i^{t} + \epsilon}$
                \STATE  $\textbf{W}'_i \leftarrow $ArgGrow$(\textbf{W}'_i, $ArgTopK$(\textbf{S}_i \cdot (\textbf{M}_i==0), k_i))$
            \ENDFOR
            \STATE  $\textbf{N}_i^{t} \leftarrow \textbf{N}_i^{t} + \textbf{M}_i$ 
        \ELSE
            \STATE $\textbf{W}'_i \leftarrow \textbf{W}'_i - \alpha \nabla(\textbf{W}'_i) \delta_t$
        \ENDIF
    
  \ENDFOR
\end{algorithmic}
\end{algorithm}

Figure~\ref{fig:dataflow} shows the training data flow of one layer using the proposed method. We use ${\textbf{W}}^t$ and ${\textbf{G}}^t$ to denote the weight and gradient tensor, respectively. $n$ is the total number of rounds of mask updates. $l^t$ is the loss to compute the gradient tensor. In the first iteration of each $\Delta T$, the weight tensor has a corresponding binary mask tensor, where zero elements are masked by 0 in the mask tensor and other elements are masked by 1. ${\textbf{N}}_t$ is the counting tensor, indicating the number of non-zero occurrences in previous mask updates.

% Figure~\ref{fig:workflow} is a toy example of a two layers convolutional neural network. The two layers have corresponding two weight tensors $W_1$ and $W_2$, two gradient tensors $G_1$ and $G_2$, two occurrence frequency counter tensors $N_1$ and $N_2$, two importance score tensors $S_1$ and $S_2$, two growth candidate tensors $C_1$ and $C_2$. The neural network is initialized at 0.5 sparsity. After initialization, following actions will be applies on the model \protect\circled{1} train model $\Delta T$ steps by updating nonzero weights; \protect\circled{2} prune 2 least magnitude weights (set them as 0) at the $\Delta T$ step; \protect\circled{3} grow 2 weights back which has the largest importance scores calculating using Equation~\ref{eq:ee}. The steps \protect\circled{1} - \protect\circled{4} will be repeated in the following training steps. 

% \subsection{Sensitivity-aware Weight Deactivation in Sparse Training}

% \vspace{-0.1cm}

\subsection{Important Weights Exploitation in Sparse Training}

% Exploitation process usually choose the best policy using visited . 

In proposed sparse training, we exploit current knowledge (weights and gradients) and define the exploitation score to help decide the mask with the highest accuracy. More specifically, we define the exploitation score $\textbf{S}_{\text{exploi}}$ in $q$-th mask update as the first item of Eq. (\ref{eq:ee}), i.e., $\textbf{S}_{\text{exploi}}=| \frac {\partial l(\textbf{W}_i^{t}, \mathcal{X})}{\partial \textbf{W}_i^{t}} |,\quad t = q \Delta T,\quad i = 1, 2, ..., L$. 

% This follows existing dynamic sparse training pipelines, which can choose a mask and get a sparse model in the neighborhood of current training state with better performance.

% To characterize how much knowledge used (the degree of exploitation), we 

% \textcolor{blue}{

% We quantify the exploitation degree when we calculate/determine the importance of the weight.

% \vspace{-1.5pt}

We further propose an evaluation metric to quantify the degree of exploitation for weight growth. With high degree of exploitation, the policy will find a model with local minimal with large loss reduction in a short time. Therefore, a growth policy is designed to have a high exploitation degree if it leads to a fast reduction in losses in the next iteration.

% We quantify the importance of a weight which exploits current knowledge (weights and gradients)~\cite{molchanov2019importance,liang2021super}. More specifically, the higher the degree, the better (greedier) model is chosen using current knowledge. Then, a better model will be generated with a larger loss reduction. A parameter is considered important if its growth leads to a significant reduction in losses in the next iteration. 

To formulate the evaluation metric, we denote $\boldsymbol{W} = [w_1^{(1,1)},w_1^{(1,2)},..., w_1^{(m_1,n_1)} ,..., w_j^{(p,q)}, ..., w_L^{(m_L,n_L)}]$ as weight of a model, where $w_j^{(p,q)}$ denotes the weight element in the $p$-th row and $q$-th column of $j$-th layer in the model. $j$-th layer has $m_j$ rows and $n_j$ columns. We further define $\boldsymbol{W}_{jpq, -jpq} = [0,...,0, w_j^{(p,q)} ,0,...,0]$ with same size of $\boldsymbol{W}$.
The degree of exploitation is denoted as $\Delta\mathcal{L}_g^{jpq}$ when the weight element in the $p$-th row and $q$-th column of $j$-th layer is grown in sparse mask update iteration, then
\begin{equation}
% \vspace{-.03in}
\begin{split}
\label{equ: sensitivity_grow}
    \Delta\mathcal{L}_g^{jpq} = \mathcal{L}(\boldsymbol{W}) - \mathcal{L}(\boldsymbol{W} + \boldsymbol{W}_{jpq,-jpq}).
\end{split}
\end{equation}

To generalize, we use $\Delta\mathcal{L}_g$ to denote the degree of exploitation of the model if k weights with indices of ${I_1, I_2, ..., I_k}$ are grown, then

\vspace{-.3in}
\begin{equation}
\begin{split}
\label{equ: sensitivity_grow_total}
    \Delta\mathcal{L}_g =  \mathcal{L}(\boldsymbol{W}) - \mathcal{L}(\boldsymbol{W} + \sum_{{n=1}}^{k}\boldsymbol{W}_{I_n,-I_n}).
\end{split}
\end{equation}

% \begin{equation}
% \begin{split}
% \label{equ: exploi}
%     S_{\text{exploi}}^{(1)} = | \textbf{W} |, \quad S_{\text{exploi}}^{(2)} = | \frac {\partial l(\textbf{W}^{t}, D)}{\partial \textbf{W}^{t}} |, \quad i = 1, 2, ..., N.
% \end{split}
% \end{equation}

% \begin{equation}
% \begin{split}
% \label{equ: exploi}
%      \quad S_{\text{exploi}} = | \frac {\partial l(\textbf{W}_i^{t}, D)}{\partial \textbf{W}_i^{t}} |,\quad t = n \Delta T,\quad i = 1, 2, ..., N
% \end{split}
% \end{equation}

% \begin{figure*}[!h]
% \begin{center}
%     %  \vspace{-.1in}
%  \includegraphics[width = 0.97\linewidth]{figures/data flow cropped.pdf}
%      \caption{Sparse training data flow{(XXX)}.}
%     %  \vspace{-3mm}
%      \label{fig:dataflow}
% \end{center}
% \end{figure*}

\subsection{Weights Coverage Exploration in Sparse Training}

Besides exploitation, we simultaneously
% prefer to 
choose masks that have never been explored so the model will not be stuck in a bad local optimum. We define our exploration score  $\textbf{S}_{\text{explor}}$ as the second item in Eq. (\ref{eq:ee}), i.e., $\textbf{S}_{\text{explor}}=\frac{\ln t}{\textbf{N}_i^{t} + \epsilon},\quad t = q \Delta T,\quad i = 1, 2, ..., L$, where $\textbf{N}_i^{t}$ is a counter tensor that collects the active (nonzero) occurrence frequency of each element. If an element with an active (nonzero) occurrence frequency of zero, it will have a corresponding higher exploration score than explored elements, thus being grown.

% If a weight has never been explored and is always zero, the mask that does not drop this weight must be unexplored. Therefore, we record the the occurrence frequency for each 1 mask and prefer mask with zero or lower frequency which correspond to unexplored or non-fully explored regions. 

% \textcolor{blue}{Applying exploration brings the benefit of increased diversity which is a popular measurement referring to differences among the explored sparse masks~\cite{vcrepinvsek2013exploration}. 
% %It is widely accepted that the balance highly depends on diversity.
% High diversity usually leads to better search performance (high accuracy), while a lack of diversity often results in stagnation (low accuracy). Numerous theoretical and empirical works have been trying to understand how diversity influence the control over exploration and exploitation~\cite{vcrepinvsek2013exploration, raghavan2018externalities, xu2014exploration}.}

% Exploration is a process that prefers visiting unexplored regions with high uncertainty. 
% That is to say, during decision making, we prefer to choose masks that have not been explored yet so that the model won't only explore the neighborhood and stuck in a local optimal. 

Inspired by RigL-ITOP~\cite{liu2021we}, we use an evaluation metric to quantify the degree of exploration for weight growth. Assume $\boldsymbol{B} = [b_{1}^{(1,1)},b_{1}^{(1,2)},..., b_{1}^{(m_1,n_1)} ,..., b_{j}^{(p,q)}, ..., b_{L}^{(m_L,n_L)}]$ is a binary vector to denote if the corresponding parameter in $\boldsymbol{W}$ is explored (1) or not (0) throughout the process of sparse training. For exploration rate~\cite{liu2021we}, we use the same formulation as RigL-ITOP~\cite{liu2021we}, i.e., $R = \frac{\sum_{{j=1}}^{L}\sum_{{p=1}}^{m_j}\sum_{{q=1}}^{n_j} b_j^{(p, q)}}{\sum_{{j=1}}^{L}{m_j}\times{n_j}}$.

\subsection{Balancing the Exploitation-Exploration Trade-off}

% \subsection{Exploitation and Exploration Theoretical Analysis}

%\textcolor{red}{@bowen}

The mask tensor search task is challenging in sparse training. Firstly, the mask search task is a high-dimensional problem due to a large number of weights in DNNs. Secondly, the search space has many local minima and saddle points \cite{han2016dsd, xie2017diverse} because of the non-convex loss function of DNNs~\cite{han2016dsd, xie2017diverse}. Therefore, the mask tensor search process is easily trapped in a bad local optimal because of its low global exploration efficiency \cite{li2015ps} or needs a longer time to fully explore the loss landscape.

%Non-convex loss functions of DNNs usually have high-dimensional parameter space and are with many local minima and saddle points \cite{xie2017diverse}. \textcolor{red}{And DNNs also contain multiple sub-networks that yield high accuracy and can be considered as local minima or saddle points in dynamic sparse training.}\textcolor{red}{(add some backup here. citation or equation)} This makes the search process trapped in one of the local optimal because of its low global exploration efficiency \cite{li2015ps} or need a longer time to fully explore the loss landscape.

A better balance between exploration and exploitation can encourage search algorithms to better understand the loss landscape and help the sparse model escape from the bad local optima. 
The importance and challenges of balancing the exploration and exploitation tradeoff have been emphasized in many studies \cite{vcrepinvsek2013exploration, wilson2021balancing}. 
%And it is even more important when dealing with hard task like high-dimensional problem and non-convex space. 
However, they have not gained enough attention in sparse training. Therefore, there is a strong need to better control the balance and we propose to consider both the exploration and exploitation scores when choosing the mask. And our importance score in Eq. (\ref{eq:ee}) combines the two scores and overcome the limitations of previous work.

\begin{table*}[!h]
% \scriptsize
% % \tiny
% \footnotesize
\small
% \addtolength{\tabcolsep}{-5pt}
% \renewcommand{\arraystretch}{1}
\centering
% \vspace{-0cm}
\resizebox{0.99\textwidth}{!}{
\begin{tabular}{l|c|ccc|ccc|}
\toprule
\textbf{Dataset} & \textbf{\#Epochs} && \textbf{CIFAR-10} &&&\textbf{CIFAR-100} & \\
% \\
% & &(393k) & (364k) & (105k) & (67k) & (8.5k) & (5.7k) & (3.7k) & (2.5k) \\
% & (\times e18) & (\times e9) & (\%) & (\times e18) & (\times e9) & (\%)  \\
\midrule
\textbf{Sparsity ratio} && 90\% & 95\% & 98\% & 90\% & 95\% & 98\% \\
\midrule
\textbf{VGG-19(Dense)} &160&  & 93.85$~\pm~$0.05 &  &  & 73.43$~\pm~$0.08 &  \\
\midrule
SNIP~\cite{lee2019snip} &160& 93.63 & 93.43 & 92.05 &72.84 & 71.83 & 58.46 \\
GraSP~\cite{wang2020picking} &160& 93.30 & 93.04 & 92.19 & 71.95 & 71.23 & 68.90 \\
SynFlow~\cite{tanaka2020pruning} &160& 93.35 & 93.45 & 92.24 & 71.77 & 71.72 & 70.94 \\
\midrule
STR~\cite{kusupati2020soft} &160& 93.73 & 93.27 & 92.21 & 71.93 & 71.14 & 69.89 \\
SIS~\cite{verma2021sparsifying} &160& 93.99 & 93.31 & 93.16 & 72.06 & 71.85 & 71.17 \\
\midrule
DeepR~\cite{bellec2017deep} &160& 90.81 & 89.59 & 86.77 & 66.83 & 63.46 & 59.58 \\
SET~\cite{mocanu2018scalable} &160& 92.46 & 91.73 & 89.18 & 72.36 & 69.81 & 65.94 \\
RigL~\cite{evci2020rigging} &160& 93.38$~\pm~$0.11 & 93.06$~\pm~$0.09 & 91.98$~\pm~$0.09 & 73.13$~\pm~$0.28 & 72.14$~\pm~$0.15 & 69.82$~\pm~$0.09 \\
DST-EE (Ours) &160& \textbf{$93.84\pm0.09$} & \textbf{$93.53\pm~0.08$} & \textbf{$92.55\pm~0.08$} & \textbf{$74.27\pm0.18$} & \textbf{$73.15\pm0.12$} & \textbf{$70.80\pm0.15$} \\
DST-EE (Ours) &250& \textbf{94.13$~\pm~$0.09} & \textbf{93.67$~\pm~$0.09} & \textbf{92.95$~\pm~$0.03} & \textbf{74.76$~\pm~$0.07} & \textbf{73.91$~\pm~$0.13} & \textbf{71.51$~\pm~$0.10} \\
\midrule
\textbf{ResNet-50(Dense)} &160&  & 94.75$~\pm~$0.01 &  &  &78.23$~\pm~$0.18  &  \\
\midrule
SNIP~\cite{lee2019snip} &160& 92.65 & 90.86 & 87.21 & 73.14 & 69.25 & 58.43 \\
GraSP~\cite{wang2020picking} &160& 92.47 & 91.32 & 88.77 & 73.28 & 70.29 & 62.12 \\
SynFlow~\cite{tanaka2020pruning} &160& 92.49 & 91.22 & 88.82 & 73.37 & 70.37 & 62.17  \\
\midrule
STR~\cite{kusupati2020soft} &160& 92.59 & 91.35 & 88.75 & 73.45 & 70.45 & 62.34  \\
SIS~\cite{verma2021sparsifying} &160& 92.81 & 91.69 & 90.11 & 73.81 & 70.62 & 62.75  \\
\midrule
RigL~\cite{evci2020rigging} &160& 94.45$~\pm~$0.43 & 93.86$~\pm~$0.25 & 93.26$~\pm~$0.22 & 76.50$~\pm~$0.33 & 76.03$~\pm~$0.34 & 75.06$~\pm~$0.27 \\

DST-EE (Ours) &160& \textbf{$94.96~\pm~0.23$} & \textbf{$94.72~\pm~0.18$} & \textbf{$94.20~\pm~0.08$} & \textbf{$78.15~\pm~0.17$} & \textbf{$77.54~\pm~0.25$} & \textbf{$75.68~\pm~0.11$} \\
DST-EE (Ours) &250& \textbf{95.01$~\pm~$0.16} & \textbf{94.92$~\pm~$0.22} & \textbf{94.53$~\pm~$0.03} & \textbf{79.16$~\pm~$0.06} & \textbf{78.66$~\pm~$0.31} & \textbf{76.38$~\pm~$0.10} \\

\bottomrule
\end{tabular}
}
\caption{Test accuracy of sparse VGG-19 and ResNet-50 on CIFAR-10/CIFAR-100 datasets. The results reported with
(mean$~\pm~$std) are run with three different random seeds. The highest test accuracy scores are marked in bold. DST-EE denotes our proposed method.}
% \vspace{0cm}
\label{tb:cifar}
\vspace{0.2cm}
\end{table*}

% \vspace{-0.2cm}

\section{Theoretical Justification}

%\subsection{Proposition 1} (Convergence)
%Inspired by \cite{ma2021effective}, 
We provide the convergence guarantee for our algorithm. We use $F(\bm{W}) = \mathbb{E}_{x\sim \mathcal{X}} f(x; W) $ to denote the loss function for our sparse training where $\mathcal{X}$ is the data generation distribution. We use $\nabla f(x;W)$ and $\nabla F(W)$ to denote the complete stochastic and accurate gradients in terms of $W$, respectively. For each round ($\Delta T$ iterations), we update the mask and use $M^{[q]}$ to denote the mask selected for the q-th round, $W^{[q]}$ to denote the model weights after $q-1$ round training. Aligned with \cite{ma2021effective}, we make the following assumptions:

\begin{assumption}
\label{assu 1}
(Smoothness). We assume the objective function $F(W)$ is partition-wise L-smooth, i.e.,
\begin{align*}
    || \nabla F(W+h) - \nabla F(W)|| \leq L ||h||,
\end{align*}
where $h$ is in the same size with $W$.
\end{assumption}

\begin{assumption}
\label{assu 2}
(Gradient noise) We assume for any t and q that
\begin{align*}
% \label{eq: ass2 2}
    \mathbb{E} [\nabla f(x_t^{(q)}&;W)] = \nabla F(W),\\
    \mathbb{E} [||\nabla f(x_t^{(q)}&;W) - \nabla F(W)||^2] \leq \sigma^2 
\end{align*}
where $\sigma > 0$ and $x_t^{(q)}$ is independent of each other.
\end{assumption}

\begin{assumption}
\label{assu 3}
(Mask-incurred error) We assume that
\begin{align*}
    ||W_t^{(q)} \odot M^{(q)} - W_t^{(q)}||^2 \leq \tau^2 ||W_t^{(q)}||^2
\end{align*}
where $\tau \in [0,1).$
\end{assumption}

%(1) loss function $F(\bm{W})$ is partition-wise L-smooth; (2) the stochastic gradient is unbiased with bounded variance; (3) the relative error introduced by mask is bounded by $\tau^2 \in [0,1)$.

Under Assumptions \ref{assu 1}-\ref{assu 3}, we establish 
%Proposition \ref{proposition: 1} to show the loss decrease in each round. Then, we get
Proposition \ref{pro: conv} to show that our sparse training algorithm converges to the stationary model at rate $O(1/\sqrt{Q})$ under the proper learning rate. 

% The proof is included in the Appendix.

%\begin{proposition}
%\label{proposition: 1} Under assumptions \ref{assu 1}-\ref{assu 3}, it holds for each $q$:
%\begin{align}
%    \mathbb{E} [F(W^{[q+1]}&)]  \leq  \mathbb{E}  [F(W^{[q]})] - \frac{\alpha \Delta T}{12} \mathbb{E} || \nabla F(W^{[q]}) ||^2 \nonumber\\
%     + &\frac{\alpha^2 L \sigma^2 \Delta T^2}{2} + \frac{2\alpha L^2 \tau^2 \Delta T^2}{3} \mathbb{E}|| W^{[q]} ||^2.
%\end{align}
%where $\alpha$ is the learning rate.
%\end{proposition}

%For Proposition \ref{proposition: 1}, we make the following remark

\begin{proposition}
\label{pro: conv}
If the learning rate $\alpha=1/(16L\Delta T\sqrt{Q})$, the sparse models generated by our algorithm after Q mask updates will converge as follows:
\begin{align}
\label{eq: pro1}
    &\frac{1}{Q} \sum_{q=1}^Q \mathbb{E}|| \nabla  F(W^{[q]} \odot M^{[q]}) ||^2 \\ =  \nonumber 
    &O\bigg{(}\frac{G}{\sqrt{Q}} + \frac{\tau^2}{Q} \sum_{q=1}^Q \mathbb{E}|| W^{[q]}||^2\bigg{)}
\end{align}
where $G$ is a constant depending on the stochastic gradient noise and the model initialization.
\end{proposition}

In regard to Proposition \ref{pro: conv}, we make the following remarks:
\begin{remark}
During dense training, we do not have error introduced by the mask and have $\tau^2=0$. As shown in Eq. (\ref{eq: pro1}), we will have $\mathbb{E}(\nabla ||F(W^{[Q]}\odot M^{[Q]}))|| \rightarrow 0$, indicating that DST-EE will converge to a stationary point as $Q\rightarrow \infty$. 
\end{remark}
\begin{remark}
%During sparse training, the model's performance is directly influenced by the error $G$ related to stochastic gradient noise and the error $\tau^2$ introduced by the mask. Our algorithm can improve mask search through a better balance between exploitation and exploration, which leads to a more accurate model.
During sparse training, the performance of the model is affected by the error $G$ associated with stochastic gradient and $\tau^2$ introduced by the mask. Our algorithm improves the mask search by a better balance between exploitation and exploration, resulting in a more accurate model.
\end{remark}

\begin{table*}[!h]
\small
\centering
\resizebox{0.99\textwidth}{!}{
\begin{tabular}{l|c|ccc|ccc|}
\toprule
\multirow{2}*{\textbf{Methods}}&\multirow{2}*{\textbf{Epochs}} & Training FLOPS &  Inference FLOPS & Top-1 Acc & Training FLOPS &  Inference FLOPS & Top-1 Acc \\
&& ($\times$ e18) & ($\times$ e9) & (\%) & ($\times$ e18) & ($\times$ e9) & (\%)  \\
\midrule
\textbf{Dense} &100&3.2 &8.2 &76.8$~\pm~$0.09 &3.2 &8.2 &76.8$~\pm~$0.09   \\
\midrule
\textbf{Sparsity ratio}      &-& & $80\%$ & & & $90\%$ &   \\
\midrule
SNIP~\cite{lee2019snip}      &-&0.23$\times$ &0.23$\times$  & -&0.10$\times$ &0.10$\times$ &-   \\
GraSP~\cite{wang2020picking}     &150&0.23$\times$ &0.23$\times$ &72.1 &0.10$\times$ &0.10$\times$ &68.1   \\
DeepR~\cite{bellec2017deep}      &-&n/a &n/a &71.7 &n/a &n/a &70.2   \\
SNFS~\cite{dettmers2019sparse}     &-&n/a &n/a &73.8 &n/a &n/a &72.3   \\
DSR~\cite{mostafa2019parameter}      &-&0.40$\times$ &0.40$\times$ &73.3 &0.30$\times$ &0.30$\times$ &71.6   \\
SET~\cite{mocanu2018scalable}      &-&0.23$\times$ &0.23$\times$ &72.9$~\pm~$0.39 &0.10$\times$ &0.10$\times$ &69.6$~\pm~$0.23   \\
RigL~\cite{evci2020rigging}      &100&0.23$\times$ &0.23$\times$ &74.6$~\pm~$0.06 &0.10$\times$ &0.10$\times$ &72.0$~\pm~$0.05  \\
MEST~\cite{yuan2021mest}      &100&0.23$\times$ &0.23$\times$ &75.39 &0.10$\times$ &0.10$\times$ &72.58   \\
RigL-ITOP~\cite{liu2021we}      &100&0.42$\times$ &0.42$\times$ &75.84$~\pm~$0.05 &0.25$\times$ &0.24$\times$ &73.82$~\pm~$0.08  \\
DST-EE(Ours)      &100&0.23$\times$ &0.42$\times$ &\textbf{76.25}$~\pm~$0.09 &0.10$\times$ &0.24$\times$ &\textbf{75.3}$~\pm~$0.06  \\
\bottomrule
\end{tabular}}
\caption{Performance of ResNet-50 on ImageNet dataset. The results reported with (mean$~\pm~$std) are run with three different  seeds.}
\vspace{-0.5cm}
\label{tb:imagenet}
\vspace{0.2cm}
\end{table*}

\section{Experimental Results}
% \textcolor{red}{(prune-and-grow vs. grow-and-prune)}

\subsection{Experimental Setup}
% \noindent\textbf{Datasets.}
% For the proposed sparse progressive distillation, 
We evaluate VGG-19 and ResNet-50 on CIFAR-10/CIFAR-100 and evaluate ResNet-50 on ImageNet. The model training and evaluation are performed with CUDA 11.1 on 8 Quadro RTX6000 GPUs and Intel(R) Xeon(R) Gold 6244 @ 3.60GHz CPU. 
% For CIFAR-10/CIFAR-100, we set the total number of training epochs as 160, while for ImageNet we set the total number of training epochs as 100. 
We use a cosine annealing learning rate scheduler with an SGD optimizer. For CIFAR-10/100, we use a batch size of 128 and set the initial learning rate to 0.1. For ImageNet, we use a batch size of 128.
% Our learning rate is scheduled with a linear warm-up for 5 epochs before reaching the initial learning rate value of 0.1. 
We use the same sparsity initialization method ERK in the state-of-the-art sparse training method such as RigL~\cite{evci2020rigging} and ITOP~\cite{liu2021we}. To further validate the generalizability of the proposed method, we conduct experiments on graph neural network for link prediction tasks on ia-email~\cite{Rossi2015TheND} and wiki-talk~\cite{Cunningham2019CreatorGI}  datasets.

% We adopt a uniform sparsity ratio across all the CONV layers while keeping the first layer dense. The other reference works (except SET [14] and RigL [11] that use uniform sparsity) use
% non-uniform sparsity, which leads to a higher computation FLOPs compared to the uniform sparsity under the same sparsity ratio. An ablation study of using hybrid sparsity schemes and non-uniform
% sparsity ratio on MEST is shown in Appendix K. The hyper-parameter setting for elastic mutation are provided in Appendix G.

% Effect of tradeoff coefficient $c$ on VGG-19.

% \vspace{-0.2cm}

\subsection{Experimental Results}
% \subsection{Accuracy Analysis}
\textbf{CIFAR-10/CIFAR-100.} The results of CIFAR-10/100 are shown in Table~\ref{tb:cifar}. We compare our method with SOTA on VGG-19 and ResNet-50 models at sparsity of 90\%, 95\%, and 98\%. To demonstrate the effectiveness of the proposed method, we compare it with three types of methods (i.e., pruning-at-initialization (SNIP, GraSP, SynFlow), dense-to-sparse training (STR, SIS), and dynamic sparse training (DeepR, SET, RigL)) from top to bottom. The results of baselines are obtained from the GraNet~\cite{liu2021sparse} paper. Overall, both pruning-at-initialization and dense to sparse methods have higher accuracy than dynamic sparse training (except for RigL (using ITOP~\cite{liu2021we} setting)). Among the various sparsity ratios, the proposed method achieves the highest accuracy for both VGG-19 and ResNet-50. Using typical training time (total training epochs is 160), there is almost no accuracy loss compared to the dense model at sparsity of 90\% on both CIFAR-10 and CIFAR-100. On both VGG-19 and ResNet-50, the proposed method has the highest accuracy compared with SOTA sparse training methods at different sparsity on both CIFAR-10 and CIFAR-10 datasets. For VGG-19, our method has up to 3.3\%, 4.6\% and 6.7\% increase in accuracy on CIFAR-10 and up to 11.1\%, 15.3\% and 18.8\% higher performance in accuracy on CIFAR-100, at sparsity ratios 90\%, 95\% and 98\%, respectively. For ResNet-50, our proposed method has accuracy improvement than RigL with the same training epochs. More specifically, on CIFAR-10, our method has 0.51, 0.86, 0.94 higher accuracy score at sparsity ratio 90\%, 95\%, 98\%, respectively. On CIFAR-100, the accuracy improvements of the proposed method compared to the SOTA sparse training method are 2.2\%, 2.0\%, 0.83\% at sparsity ratios of 90\%, 95\%, and 98\%, respectively.

\noindent\textbf{ImageNet.} Table~\ref{tb:imagenet} shows the top-1 accuracy results, training and inference FLOPS on ResNet50 / ImageNet. 
% We compare FDST with three types of SOTA methods same as CIFAR-10/100 datasets (i.e., pruning at initilization, dense-to-sparse training and dynamic sparse training). 
We use the dense training model as our baseline. For other baselines, we select SNIP~\cite{lee2019snip} and GraSP~\cite{wang2020picking} as the static mask training baselines while adopting DeepR~\cite{bellec2017deep}, SNFS~\cite{dettmers2019sparse}, DSR~\cite{mostafa2019parameter}, SET~\cite{mocanu2018scalable}, RigL~\cite{evci2020rigging}, MEST~\cite{yuan2021mest}, RigL-ITOP~\cite{liu2021we} as the dynamic mask training baselines as shown in Table~\ref{tb:imagenet}. Compared to static mask training baselines, our proposed method has up to 5.8\% and 10.6\% increase in accuracy. For the dynamic mask training baselines, RigL is the recently popular baseline, compared with which the proposed method has 2.2\% and 3.7\% higher Top-1 accuracy at sparsity ratios of 80\% and 90\%, respectively. For the other two better baselines of sparse training, MEST and RigL-ITOP, our method has 1.1\% and 0.5\% higher accuracy at a sparsity ratio of 0.8, and  3.7\% and 1.48\% accuracy improvement at a sparsity ratio of 0.9, respectively.
% Show accuracy curve with respect to training epoches on CIFAR-10, CIFAR-100, ImageNet.

% \begin{figure}[!h]
% %\vspace{-.1in}
% % \begin{center}
% \includegraphics[width=0.99\linewidth]{figures/imagenet_acc_v1.pdf}
% \caption{Comparing the Top-1 accuracy of ResNet at different sparsity ratios on ImageNet}
% %  \vspace{-3mm}
% \label{fig:imagenet}
% % \end{center}
% \end{figure}

% \textbf{Faster convergence.}

% \begin{figure}[!h]
%     \centering
%     % \vspace{-.25in}
%     \subfloat[Exploration degree on VGG-19 / CIFAR-100 with sparsity=0.90.]{\label{fig:motivation1}
%     \hspace{-.0in}\includegraphics[width=0.465\columnwidth]{figures/vgg19_sparsity_0.9_exploration_degree.pdf}}
%     %\caption{fig1}
%     % \vspace{-.0in}
%     \quad
%     \subfloat[Test accuracy of VGG-19 / CIFAR-100 with sparsity=0.90.]{\label{fig:motivation2}
%     \hspace{-.0in}\includegraphics[width=0.465\columnwidth]{figures/vgg19_sparsity_0.9_acc.pdf}}
%     \caption{(a)  (b) 
%     }
%     \label{fig:exploration_degree}
%     % \vspace{0.2in}
% \end{figure}

\begin{table}[!h]
\small
\centering
\resizebox{1.0\columnwidth}{!}{
\begin{tabular}{l|c|ccc}
\toprule
\multirow{2}*{\textbf{Methods}}&\multirow{2}*{\textbf{Epochs}} &  Sparsity ratio &  Sparsity ratio &  Sparsity ratio\\
&& 80\% &  90\% &  98\%  \\
\midrule
\textbf{Dense} &- & &79.72 & \\
\midrule
\textbf{Prune-from-dense}      &60& 79.05&78.34&78.08  \\
\textbf{DST-EE (ours)}      &50& \textbf{79.28}&\textbf{79.13}&\textbf{78.58}    \\
\bottomrule
\end{tabular}}
\caption{GNN link prediction Results tasks on wiki-talk~\cite{Cunningham2019CreatorGI}.}
\label{tb:wiki-talk}
\vspace{0.2cm}
\end{table}

\begin{table}[!h]
\small
\centering
\resizebox{1.0\columnwidth}{!}{
\begin{tabular}{l|c|ccc}
\toprule
\multirow{2}*{\textbf{Methods}}&\multirow{2}*{\textbf{Epochs}} &  Sparsity ratio &  Sparsity ratio &  Sparsity ratio\\
&& 80\% &  90\% &  98\%  \\
\midrule
\textbf{Dense} &- & &83.47 & \\
\midrule
\textbf{Prune-from-dense}      &60& 83.19&82.95&67.18  \\
\textbf{DST-EE (ours)}      &50& \textbf{83.77}&\textbf{83.29}&\textbf{82.82}   \\
\bottomrule
\end{tabular}}
\caption{GNN link prediction results on ia-email~\cite{Rossi2015TheND}.}
\label{tb:ia-email}
\vspace{0.2cm}
\end{table}

\begin{figure}
    \centering
    % \vspace{-.25in}
    \subfloat[CIFAR-100 / Sparsity=0.95]{\label{fig:explore_d}
    \hspace{-.0in}\includegraphics[width=.23\textwidth]{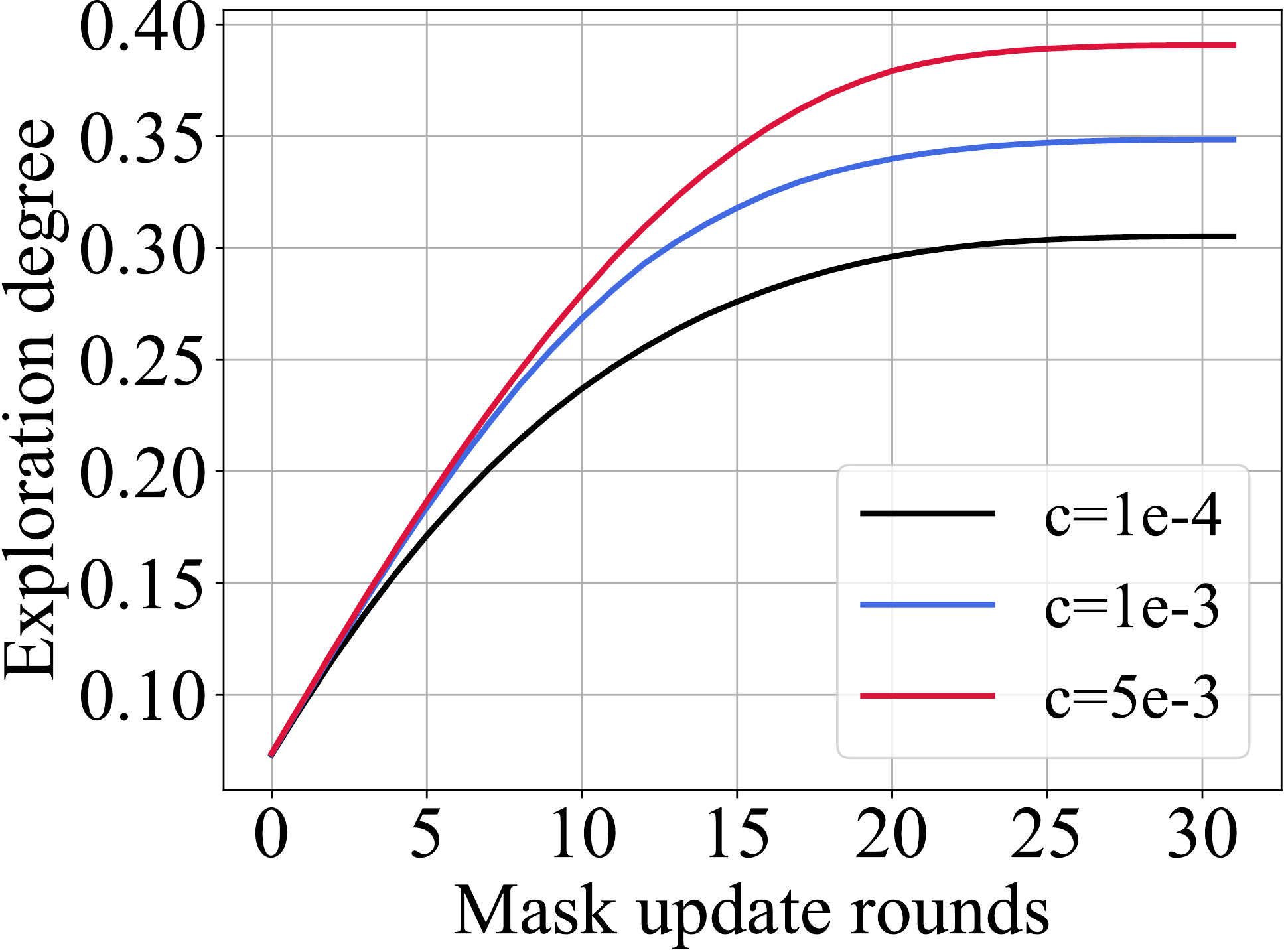}
    \hfill
    \includegraphics[width=.23\textwidth]{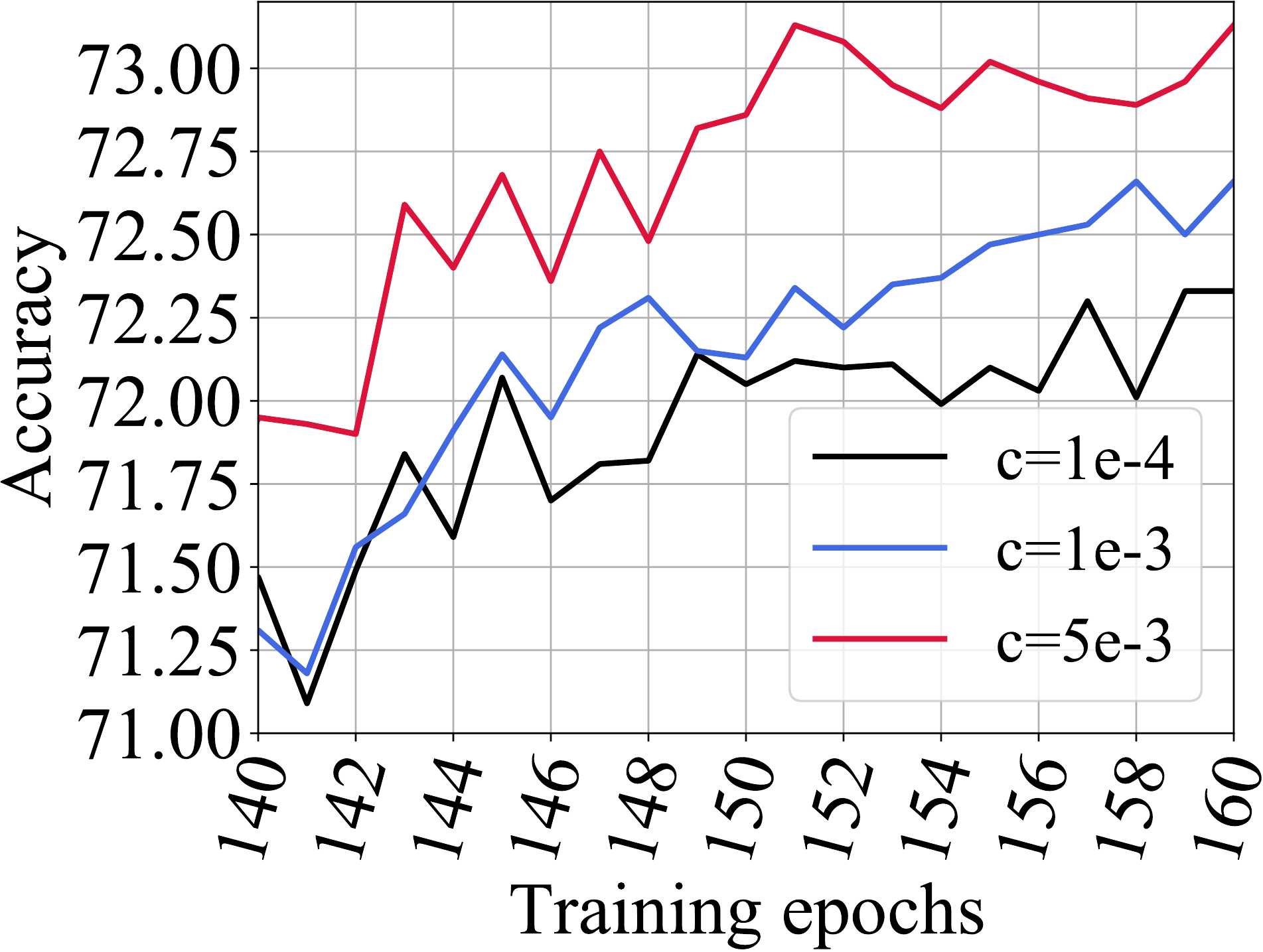}}
    \quad
    \subfloat[CIFAR-10 / Sparsity=0.95]{\label{fig:explore_a}
    \hspace{-.0in}\includegraphics[width=.23\textwidth]{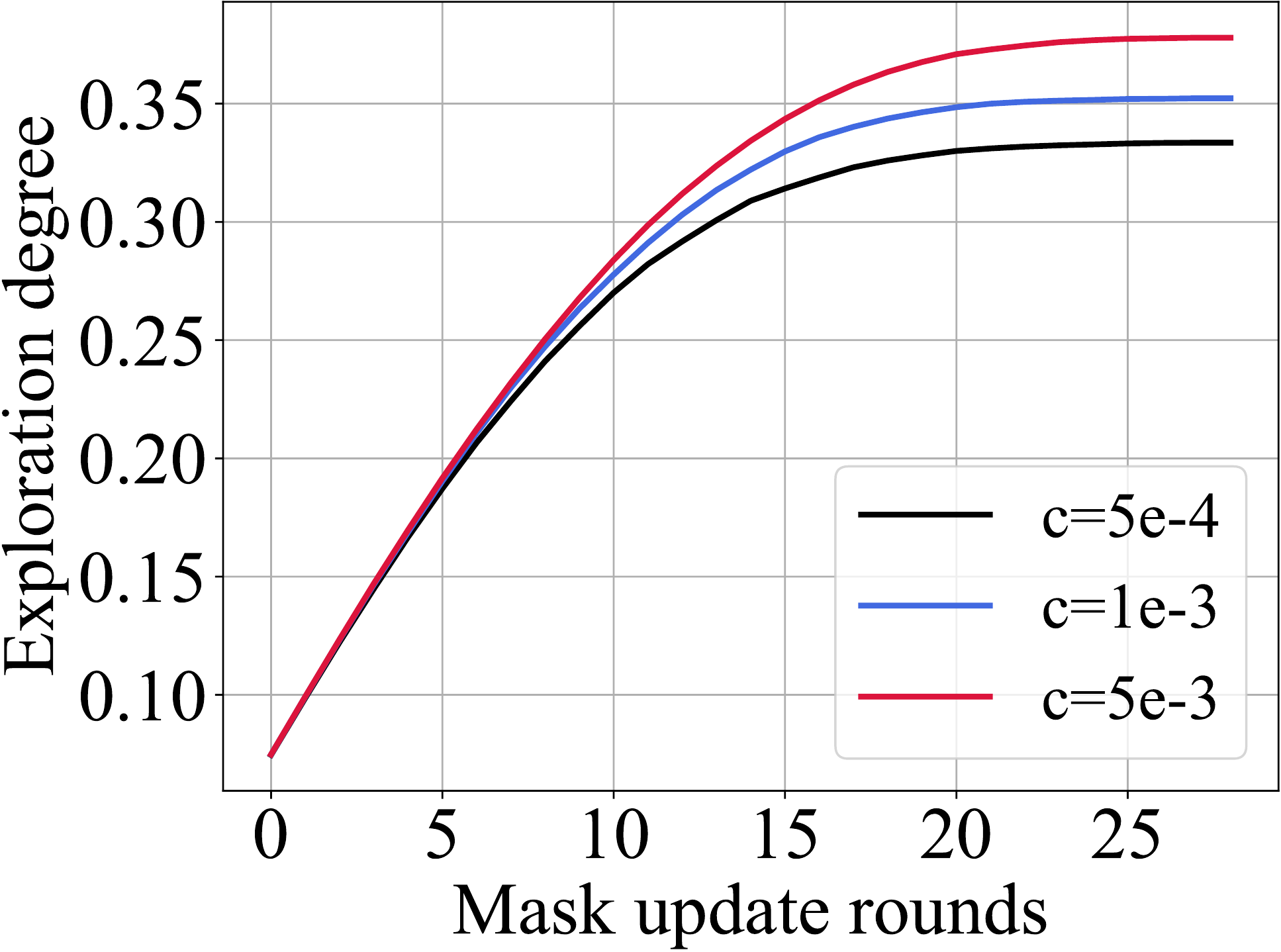}
    \hfill
    \includegraphics[width=.23\textwidth]{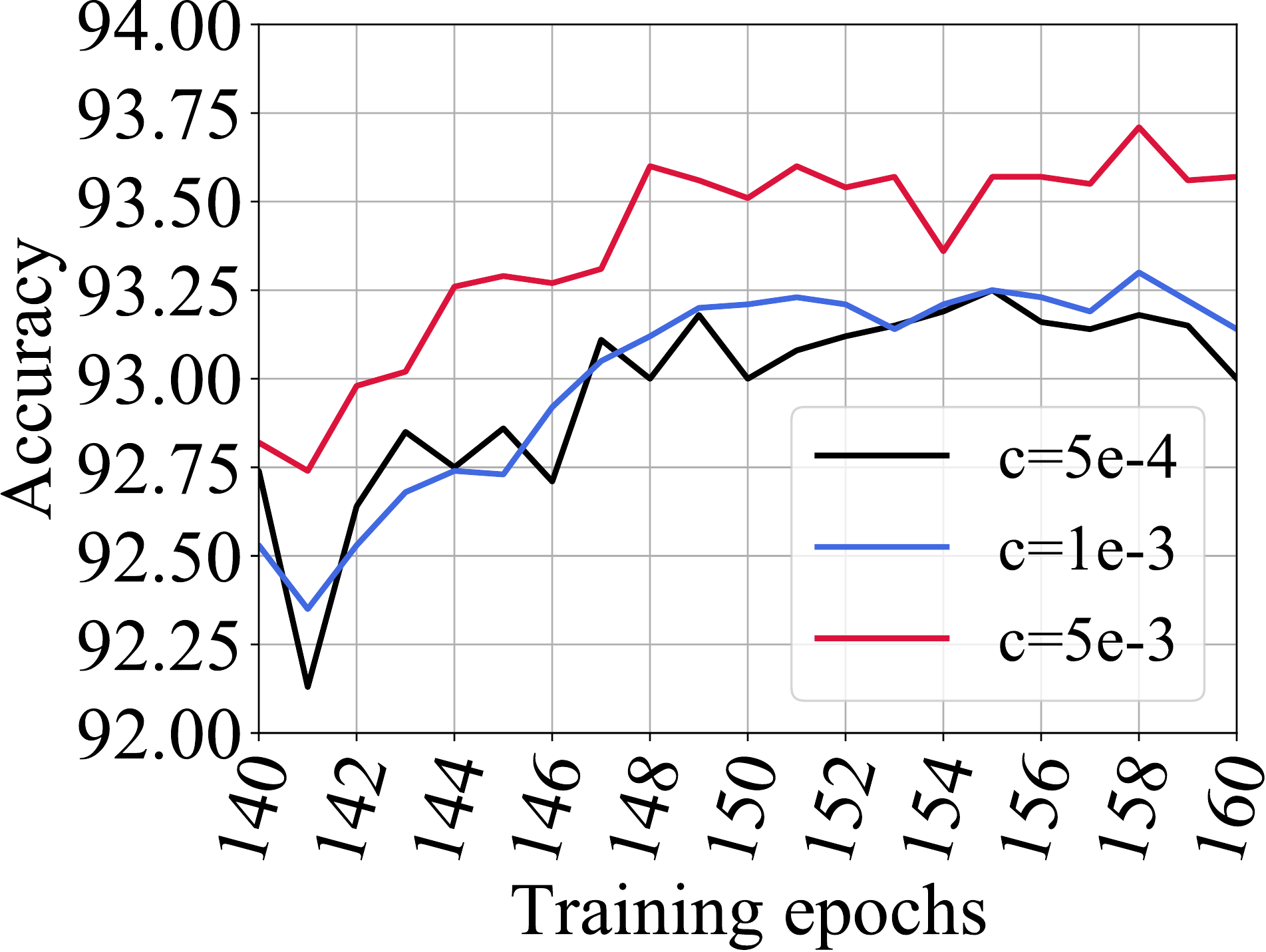}}
    \caption{The figure shows the relation of exploration degrees and test accuracy on CIFAR-10 and CIFAR-100 with a sparsity 0.95.
    }
    \label{fig:exploration_degree}
    % \vspace{0.2in}
\end{figure}

\noindent\textbf{Graph Neural Network.}
Experimental results of sparse training of graph neural network on wiki-talk~\cite{Cunningham2019CreatorGI} and ia-email~\cite{Rossi2015TheND} for link prediction task are shown in Table~\ref{tb:wiki-talk} and Table~\ref{tb:ia-email}, respectively. We apply the proposed method to the two fully connected layers with uniform sparsity ratios at different sparsity levels, which are 80\%, 90\%, and 98\%. We report the prediction accuracy of the best model searched in 50 training epochs. We compare our method with both the dense model and the best sparse model pruned from the dense model using ADMM algorithm. The prune-from-dense models are trained for 60 epochs in total, which of 20 pretraining epochs, 20 reweighted training epochs, and 20 retraining epochs after pruning. Experimental results show that at a sparsity of 0.8, our sparse training method has even better accuracy than the dense model. The proposed method has accuracy improvement compared with prune-from-dense on both datasets using even fewer training epochs. On wiki-talk~\cite{Cunningham2019CreatorGI}, our method has 0.29\%, 1.0\% and 0.64\% higher accuracy than prune-from-dense using ADMM algorithm at sparsity ratios of 80\%, 90\% and 98\%, respectively. On ia-email~\cite{Rossi2015TheND}, the proposed method has up to 23.3\% accuracy improvement than prune-from-dense at a sparsity ratio of 98\%. 
% The similar promising results on GNNs justify the generalizability of our proposed methods.

% \subsection{Design Exploration on Different Exploration Degrees}

% \subsection{Effects of Different Learning Rates.}
% \subsubsection{Effects of Extended Training Time.} We test our method using extended training time (total number of training epochs is 250) on VGG-19/ResNet-50 models and  CIFAR-10/CIFAR-100 datasets. On CIFAR-10 (90\% sparsity ratio), and CIFAR-100 (90\% and 95\% sparsity ratios), there is no decrease in the accuracy of both models compared to dense models. For VGG-19, there is only 0.19\% and 0.96\% accuracy degradation on CIFAR-10 at 95\% and 98\% sparsity ratios compared to the dense model, while on CIFAR-100, DST-EE outperforms SOTA methods by at least 1.7\%, 2.0\% and 2.5\% at 90\%, 95\%, and 98\% sparsity ratios, respectively. For ResNet-50, there is no accuracy drop at 90\% and 95\% sparsity ratios on both datasets. On CIFAR-10, it has 8.4\%, 6.5\%, 6.4\%, 6.5\%, 4.9\% and 1.4\% higher accuracy performance at 98\% sparsity ratio compared to SNIP~\cite{lee2019snip}, GraSP~\cite{wang2020picking}, SynFlow~\cite{tanaka2020pruning}, STR~\cite{kusupati2020soft}, SIS~\cite{verma2021sparsifying} and RigL~\cite{evci2020rigging}, respectively. On CIFAR-100, DST-EE has at least 1.8\% and at most 30.7\% accuracy improvement compared to SOTA.

\subsection{Design Exploration on Different Exploration Degrees.} We investigate the effect of coefficients on exploration degree and test accuracy on VGG-19, CIFAR-10 / CIFAR-100 datasets as shown in Figure~\ref{fig:exploration_degree}. The left subfigure in Figure~\ref{fig:explore_d} shows the different exploration degree curves generated using different tradeoff coefficients on CIFAR-100 with a sparsity of 0.95. We could see the larger $c$, the higher degree of exploration of the sparse model. The right subfigure in Figure~\ref{fig:explore_d} illustrates the test accuracy curves for different coefficients. Within the coefficient range, the larger $c$, the higher test accuracy. The combination of these two subfigures unveils the observation that the higher the exploration degree or higher weights coverage, the higher the test accuracy score. Similar observations are shown in  Figure~\ref{fig:explore_a}, which validate our methods.

\section{Conclusion}
In this paper, we propose important weights exploitation and coverage exploration-driven growth strategy to characterize and assist explainable sparse training, update the sparse masks and search for the “best possible” subnetwork.  We provide theoretical analysis for the proposed exploitation and exploration method and clarify its convergence property. 
% In particular, we justify the influence of the mask-induced error.
We further provide the quantitative analysis of the strategy and show the advantage of the proposed method.
% Different from existing greedy-based method, which only exploit the current knowledge, we preferentially grow weights that have never been covered before, 
% resulting in a higher degree of exploration in the same training time, therefore, higher test accuracy.
We design the acquisition function to evaluate the importance of non-active weights for growth and grow the weights with top-k highest importance scores, considering the balance between exploitation and exploration. Extensive experiments on various deep learning tasks on both convolutional neural networks and graph neural networks show the advantage of DST-EE over existing sparse training methods. 
We conduct experiments to quantitatively analyze the effects of exploration degree. The observations validate the proposed method, i.e., our method could achieve a higher exploration degree and thus a higher test accuracy compared to greedy-based methods.

\section*{Acknowledgement}
This work was partially funded by the Semiconductor Research Corporation (SRC) Artificial Intelligence Hardware program,  and the UIUC HACC program.

% \section{Citations}

% Some examples of references. A paginated journal article~\cite{Abril07}, an enumerated journal article~\cite{Cohen07}, a reference to an entire issue~\cite{JCohen96}, a monograph (whole book) ~\cite{Kosiur01}, a monograph/whole book in a series (see 2a in spec. document)~\cite{Harel79}, a divisible-book such as an anthology or compilation~\cite{Editor00} followed by the same example, however we only output the series if the volume number is given~\cite{Editor00a} (so Editor00a's series should NOT be present since it has no vol. no.), a chapter in a divisible book~\cite{Spector90}, a chapter in a divisible book in a series~\cite{Douglass98}, a multi-volume work as book~\cite{Knuth97}, an article in a proceedings (of a conference, symposium, workshop for example) (paginated proceedings article)~\cite{Andler79}, a proceedings article with all possible elements~\cite{Smith10}, an example of an enumerated proceedings article~\cite{VanGundy07}, an informally published work~\cite{Harel78}, a doctoral dissertation~\cite{Clarkson85}, a master's thesis~\cite{anisi03}, an finally two online documents or world wide web resources~\cite{Thornburg01, Ablamowicz07}.

% \begin{acks}
%  This work was supported by the [...] Research Fund of [...] (Number [...]). Additional funding was provided by [...] and [...]. We also thank [...] for contributing [...].
% \end{acks}

%\clearpage

% \bibliographystyle{ACM-Reference-Format}

% \clearpage
\vspace{-0.5pt}
\scriptsize
\bibliography{dac2023}

\end{document}